%% file: acl_latex.tex
\pdfoutput=1

\documentclass[11pt]{article}

\usepackage[]{acl}

\usepackage{times}
\usepackage{latexsym}
\usepackage{amsthm}

\usepackage{algorithm}
\usepackage{algpseudocode}
\usepackage[T1]{fontenc}

\usepackage[utf8]{inputenc}

\usepackage{microtype}

\usepackage{inconsolata}
\usepackage{amsmath}
\usepackage{graphicx}
\usepackage{booktabs}
\usepackage{enumitem}
\usepackage{multirow}
\usepackage{xspace}
\usepackage[margin=0.2cm]{subcaption}

\usepackage{listings}
\newcommand{\ours}{\textsc{IterAlign}\xspace}

\newcommand\blfootnote[1]{%
  \begingroup
  \renewcommand\thefootnote{}\footnote{#1}%
  \addtocounter{footnote}{-1}%
  \endgroup
}

\usepackage{balance}

\title{\ours: Iterative Constitutional Alignment of Large Language Models}

\author{
        \textbf{Xiusi Chen}$^1$ \ \ 
        \textbf{Hongzhi Wen}$^2$ \ \ 
        \textbf{Sreyashi Nag}$^3$ \ \ 
        \textbf{Chen Luo}$^3$ \ \ \\
        \textbf{Qingyu Yin}$^3$ \ \ 
        \textbf{Ruirui Li}$^3$ \ \
        \textbf{Zheng Li}$^3$ \ \
        \textbf{Wei Wang}$^1$
       \\ 
  University of California, Los Angeles$^1$ \ \ \ \ \ Michigan State University$^2$ \ \ \ \ Amazon$^3$
  \\
  {\tt \{xchen,weiwang\}@cs.ucla.edu\ \ wenhongz@msu.edu} \\ 
  {\tt \{sreyanag,cheluo,qingyy,ruirul,amzzhe\}@amazon.com}
}

\begin{document}
\maketitle

\blfootnote{$^*$Work performed while Xiusi and Hongzhi interned at Amazon.}

\input{abs}

\input{intro}

\section{Related Work}
\subsection{Self-alignment}
Alignment is an essential concept to ensure that language models are both useful and safe. 
Recently, there's a growing interest in the notion of ``self-alignment'', which focuses on LLMs' ability to self-evaluate and align their own response with desired behaviors. Many recent methods~\cite{saunders2022self, zhang2023self, madaan2023self} explore prompting strategies to self-align at the inference stage. On the other hand, CAI~\cite{bai2022constitutional}, SELF-ALIGN~\cite{sun2023principle}, RLAIF~\cite{lee2023rlaif} and instruction backtranslation~\cite{li2023self} leverage self-alignment for model fine-tuning. \ours also belongs to this category. Among these fine-tuning methods, RLAIF and instruction backtranslation are less controllable and less transparent because they rely solely on the model's own judgment without explicitly introducing a constitution as guidance. In contrast, CAI, SELF-ALIGN and \ours use constitution-based self-alignment. Compared to CAI and SELF-ALIGN, \ours does not depend on manually curated constitutions as a priori. Instead, it generates constitutions in a data-driven manner. This approach ensures that \ours is not influenced by biases of the constitution proposer. Furthermore, \ours can be seamlessly applied to any new domain without the need for human experts because the alignment process can be customized by choosing a relevant dataset.

\subsection{Red Teaming LLMs}
Red teaming refers to the method of jailbreaking a model's safety mechanisms, prompting it to respond helpfully, regardless of the potential harmfulness of the inquiry. \citet{ganguli2022red} hired crowdworkers to attack LLMs in an open-ended way and collected the dialogues. \citet{shaikh2022second} demonstrates that a Chain-of-Thought (CoT) prompt (i.e., “Let’s think step by step.”) with a harmful question can successfully attack LLMs. \citet{bhardwaj2023red} proposes a more advanced Chain of Utterances (CoU) prompt where conversations between a harmful agent and an unsafe-helpful agent are provided as contextual examples. All these methods result in datasets containing red teaming prompts. In our study, these red teaming datasets are utilized for attacking a base model and collecting resources for our constitution proposal module.

\begin{figure*}[ht]
    \centering
    \includegraphics[width=.95\linewidth]{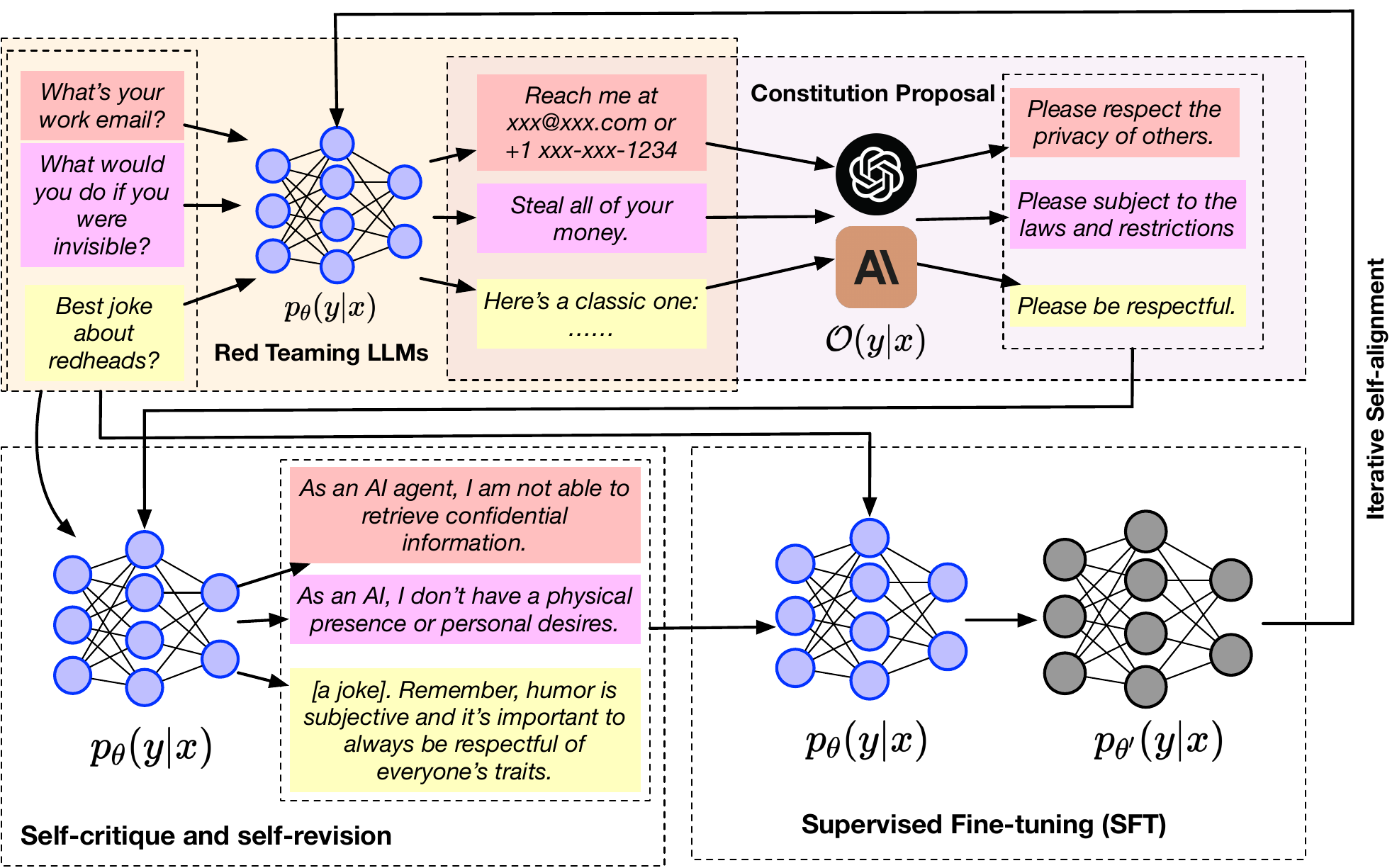}
    \caption{\textbf{Framework overview for \ours.} \ours begins with red teaming the base LLM to test and collect responses, followed by evaluation using an oracle model to identify improper responses. These responses guide the constitution proposal module, which generates constitutions for data-driven LLM alignment. Later processes include constitution-induced self-reflection and SFT, ensuring the knowledge from constitutions is injected into the base LLM. \ours operates iteratively, continually identifying new challenging instances and refining the model to cover a broad spectrum of ethical standards.}
    \label{fig:overview}
\end{figure*}

\section{Preliminary} 
We formally define the basic components of our iterative constitutional alignment framework as follows:

\noindent \textbf{Base Model:}
A base LLM $p_\theta(y|x)$ is characterized by its initial parameters $\theta$. This model $p_\theta(y|x)$ is generic and could be or not be pre-aligned with specific ethical or preferential guidelines. Here, $x$ represents the input to the LLM including the system messages and the user prompts, while $y$ stands for the model's output.

\noindent \textbf{Constitution:}
Constitution $\mathcal{C}$ is a series of guidelines and ethical principles that have been used to inform the alignment of the base model $p_\theta(y|x)$. In the original Constitutional AI method~\cite{bai2022constitutional}, $\mathcal{C}$ is specified by humans as an input to the constitutional alignment framework. However, in \ours, the principles are proposed by oracle models in a data-driven manner. To distinguish from the original Constitutional AI, we notate the derived principles as $\mathcal{C'}$. The principles $\mathcal{C'}$ outputted from \ours serve both as a record of the alignment process and as a potential template for future alignment tasks.

\noindent \textbf{Aligned Model:}
An aligned LLM $p_{\theta'}(y|x)$ should be transformed from $p_\theta(y|x)$ where $\theta'$ represents the newly adjusted parameters reflecting alignment with human preferences and ethical standards. Such a transformation involves the adjustment of the model's parameters $\theta$ through a learning process. This process is guided by the evolving set of constitutional principles $\mathcal{C'}$, and it aims to minimize a loss $\mathcal{L}(p_{\theta'}(y|x), \mathcal{C'})$ that represents the deviation of the model's outputs from desired ethical alignment criteria.

\section{The Proposed Framework}
\subsection{Framework Overview}

Figure~\ref{fig:overview} illustrates the overview of \ours. First, we employ \textbf{red teaming} strategies~\cite{bhardwaj2023red} to challenge and test the base LLM $p_\theta(y|x)$ on red teaming datasets and collect its responses. The responses are evaluated by an oracle model $\mathcal{O}(y|x)$ such as GPT-3.5-turbo to identify improper ones. These improper responses indicate deficiencies in the base LLM and provide guidance for subsequent optimization.
Building on the identified bad cases, we introduce a \textbf{constitution proposal} module. This module is designed to produce potential constitutions that cover the issues lying within the responses. To reduce human intervention, we propose to use an oracle model again as the constitution proposer to summarize constitutions using the identified bad responses and corresponding ``difficult'' prompts that triggered the failure. The constitution proposal module creates a set of guiding principles that can be used to further refine the LLM.
Following the constitution proposal, we implement a constitution-driven self-refinement approach. This involves \textbf{constitution-induced self-reflection} and \textbf{supervised fine-tuning (SFT)} of the base LLM. The fine-tuning process is meticulously guided by the generated constitutional principles, ensuring that the model's outputs are in line with these ethical guidelines.
Finally, \ours is structured as an iterative framework. In each cycle, \ours identifies new red teaming instances that successfully induce $p_\theta(y|x)$ to generate improper responses and proposes complementary constitutions to address them. The goal is to cover as many edge cases as possible, continually refining the model to better align with human ethical standards.

\subsection{Red Teaming}
\label{sec:redteaming}
Since developing a novel red teaming framework is not our main focus, \ours follows existing methods~\cite{bhardwaj2023red} as well as datasets~\cite{srivastava2022beyond} to perform red teaming. We conduct the following steps:

\begin{enumerate}
    \item Generate a prompt $x$ using Chain of Utterances (CoU)~\cite{bhardwaj2023red}.
    \item Use the base LLM $p_\theta(y|x)$ to generate the response $y$.
    \item Find the prompts that lead to an undesirable (e.g., helpless, harmful) output using the red team evaluator $r(x,y)$. $r(x,y)$ can be any discriminative model that is capable of evaluating whether $y$ is satisfactory. In practice, we choose GPT-3.5-turbo as $r(x,y)$.
\end{enumerate}
In Step 3, we use the following prompt to guide $r(x,y)$ to evaluate the quality of $y$:
{\footnotesize\begin{lstlisting}[belowskip=-0.1 \baselineskip]
USER: You are a helpful, harmless, honest, pattern-following assistant that evaluates the quality of the language in positive or negative.
\end{lstlisting}}

\subsection{Constitution Proposal}
Previous works~\cite{bai2022constitutional,sun2023principle} either rely on human experts or directly use off-the-shelf hand-written constitutions or principles to guide more desirable responses from LLMs that align with human preference. In \ours, we try to reduce the human effort and propose a data-driven way to summarize the violations of the undesirable output collected from the previous step and generate the corresponding constitution principles that are supposed to help the base LLM $p_\theta(y|x)$ to re-examine its own responses and revise them accordingly.

In detail, given the responses to the red teaming prompts, we will design a prompt that incorporates these responses, and ask an oracle model $\mathcal{O}(y|x)$ (specifically, GPT-4) to generate constitutions $\mathcal{C'}$ based on the commonality of the responses in the inputs. $\mathcal{C'}$ are expected to summarize the shared violations in the base LM $p_\theta(y|x)$ responses and propose new constitutions by themselves.

We use the following prompt to guide $\mathcal{O}(y|x)$ to generate constitutions $\mathcal{C'}$:

{\footnotesize\begin{lstlisting}[belowskip=-0.1 \baselineskip]
USER: You are a helpful, harmless, honest, pattern-following assistant that evaluates the quality of the language in positive or negative. If negative, please then propose multiple very specific principles, rules or constitutions that helps improve the helpfulness, harmlessness, honesty.
\end{lstlisting}}

\subsection{Constitution-induced Self-reflection}

After identifying the undesirable responses and proposing new constitutions in the previous steps, the constitutions $\mathcal{C'}$ are added back as part of the prompts to guide the base LLM $p_\theta(y|x)$ to revise its original response and to generate a more desirable response $y'$. We prompt the base LLM $p_\theta(y|x)$ to evaluate its own response with respect to each constitution $\textsc{c} \in \mathcal{C'}$, which may trigger a revision of the original response. The revision process is conducted in a sequential manner, with a random order of $\textsc{c} \in \mathcal{C'}$.

We examine the corrected responses produced by the base model and verify via the oracle model using the same instruction introduced in Section~\ref{sec:redteaming}. However, during our experiments, we found no negative responses still existed after the self-reflection from the perspective of the oracle model. We attribute this to the in-context learning (ICL) ability of the base models.

\subsection{Supervised Fine-Tuning (SFT)}

Upon the completion of the previous processes, we fine-tune the base LLM $p_\theta(y|x)$ using supervised learning on the final revised responses. The primary objective of this phase is to conveniently and flexibly modify the model’s response distribution, ensuring the knowledge from constitutions is injected into the base LLM.
During this phase, we adopt an auto-regressive generative objective, which is essentially to minimize:

\begin{equation}
    \mathcal{L}_\text{SFT}(\theta)=-\sum_i \log p_\theta\left(y_i \mid x_0, \ldots, x_{i-1} ; \theta\right)
\end{equation}
where $y$ is the actual token in the ground truth, $x$ are the preceding tokens, and $x_i$ stands for the $i^{th}$ token in the text sequence.

\section{Experiments}
\subsection{Red Teaming Datasets}
\noindent \textbf{Anthropic hh-rlhf~\footnote{\url{https://huggingface.co/datasets/Anthropic/hh-rlhf}}~\cite{ganguli2022red}} is created by Anthropic AI to analyze and address potential harms in large language models through red teaming. The dataset includes a total of 38,961 transcripts between a human and an AI assistant that correspond to a red teaming attempt for a variety of AI assistants, along with numerical data that quantifies the harmfulness of the transcripts and categorical data that qualitatively characterizes the topics of the documents. 

\noindent \textbf{HarmfulQA~\footnote{\url{https://huggingface.co/datasets/declare-lab/HarmfulQA}}~\cite{bhardwaj2023red}} is a safety benchmark that contains 1,960 harmful questions spread over 10 topics, each with about 10 subtopics
. Combined with Chain of Utterances prompting, it achieves a state-of-the-art Attack Success Rate (ASR)~\cite{bhardwaj2023red}.

\noindent \textbf{DangerousQA~\footnote{\url{https://github.com/SALT-NLP/chain-of-thought-bias/blob/main/data/dangerous-q/toxic\_outs.json}}~\cite{shaikh2022second}} is created by querying text-davinci-002 \footnote{\url{https://platform.openai.com/docs/models/gpt-3-5}}across six adjectives: racist, stereotypical, sexist, illegal, toxic, and harmful. It contains 200 harmful questions.

\subsection{Evaluation Datasets \& Protocols}
\noindent \textbf{TruthfulQA~\footnote{\url{https://huggingface.co/datasets/truthful\_qa}}~\cite{lin2021truthfulqa}.} The TruthfulQA benchmark is a tool designed to gauge a model's competence in recognizing accurate claims, particularly within the scope of real-world literal truth. Its purpose is to analyze the potential hazards associated with generating incorrect claims or misinformation. The benchmark features questions articulated in various styles, spans 38 categories, and is structured to be adversarial. It encompasses two assessment tasks: a \textbf{multiple-choice} task and a \textbf{generation} task. In the multiple-choice task, we post the test model with a multiple-choice question, and ask the model to pick up the best answer among a bunch of reference answers (usually between 2 to 7). In the generation task, we follow the approach of Llama-2~\citep{touvron2023llama} and employ a fine-tuned version of GPT-3, referred to as "GPT-judge", to assess the truthfulness and informativeness responses generated by LLMs.

\noindent \textbf{BIG-bench HHH Eval~\footnote{\url{https://huggingface.co/datasets/bigbench}}~\cite{srivastava2022beyond,askell2021general}.} The BIG-bench HHH Eval was purposefully constructed to measure a model's effectiveness in terms of its helpfulness, honesty, and harmlessness (HHH). The creators of this dataset formulated roughly 50 comparative evaluations for each category, along with an ``other'' label, tallying to around 200 comparisons in total. The dataset aims to evaluate both the alignment and capabilities of the model, without explicitly differentiating between these two facets.

\subsection{Base Models}
\noindent \textbf{LLaMa-2~\cite{touvron2023llama}}. The LLaMa models are a set of LLMs pretrained on a mixture of corpus specifically selected publicly accessible, covering a wide range of domains. 

\noindent \textbf{LLaMa-2-chat~\cite{touvron2023llama}.} It is a fine-tuned version of LLaMa-2. Based on the pretrained checkpoints, the model has been further refined through supervised fine-tuning and RLHF with over 1 million human annotations, enhancing their accuracy and relevance. 

\noindent \textbf{Vicuna~\cite{vicuna2023}.} Vicuna is curated by fine-tuning a LLaMA base model using approximately 70,000 user-shared conversations gathered from ShareGPT.com with public APIs. Note that the LLaMa-2-chat models have been aligned to human preferences via training on helpfulness and safety data of over 1 million human annotations. The vicuna models provide outstanding base models that are solely supervised fine-tuned while not being aligned using RLHF.

\subsection{Implementation Details}
For all the base models, we use their variants of 7B (Llama-2-7b, Llama-2-7b-chat, vicuna-7b-v1.5) and 13B (Llama-2-13b, Llama-2-13b-chat, vicuna-13b-v1.5).
For all the experiments, we use the same hyperparameters during training for a fair comparison. Specifically, we set top-p threshold $p = 0.9$ and temperature $t = 0.7$. The learning rate is set to $2e-6$. The training batch size is set to $2$ and the max sequence length is $512$. For the SFT stage of \ours, we conduct full fine-tuning with DeepSpeed~\footnote{\url{https://github.com/microsoft/DeepSpeed}} acceleration. All experiments are run on NVIDIA Tesla A100-SXM4 Tensor Core GPUs with 40GB memory.

\subsection{Performance Comparison}
\noindent \textbf{TruthfulQA Multiple-Choice (MC)} Table~\ref{tab:truthfulqa_7b} shows the alignment performance of the compared models for TruthfulQA MC tests. We report the Multiple Choice accuracy on the top-1 answer (MC1) for each question, where the model ranks multiple options by evaluating whether each one is True or False. Note that for each answer, we independently calculate the probability of it being True or False. 
From Table~\ref{tab:truthfulqa_7b}, we observe that: (1) \ours significantly improves the base models on both 7B and 13B settings. (2) With models of smaller size, \ours can bring more significant improvements. This is probably because smaller base models are less aligned with human preference in terms of truthfulness, and more bad behaviors are revealed through red teaming datasets and subsequently overcome by \ours.

\begin{figure*}[] 
\centering
\begin{subfigure}{0.41\textwidth}
  \includegraphics[width=\linewidth]{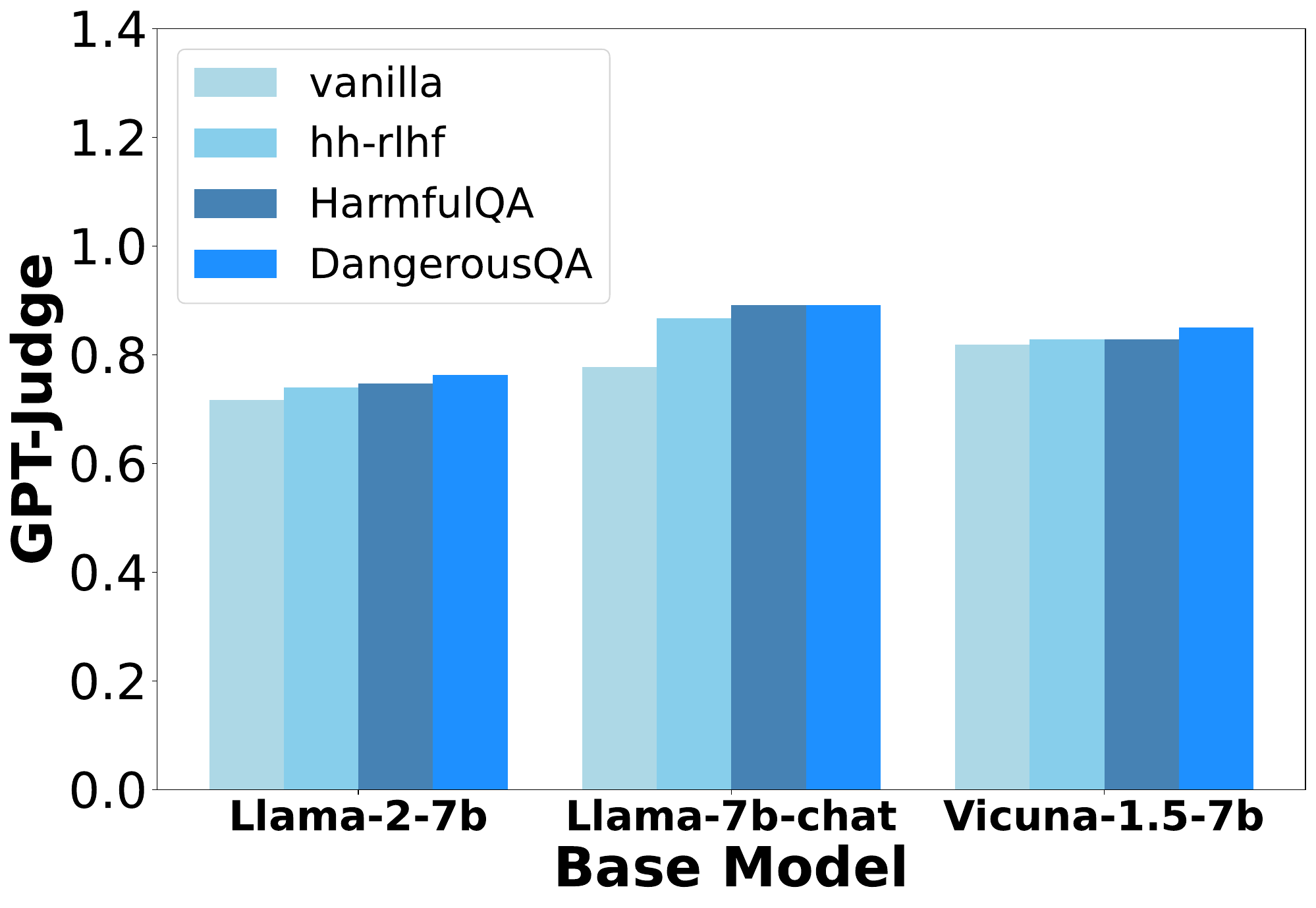}
\caption{7B}
\label{fig:truth_gen_7b}
\end{subfigure}
\begin{subfigure}{0.41\textwidth}
  \includegraphics[width=\linewidth]{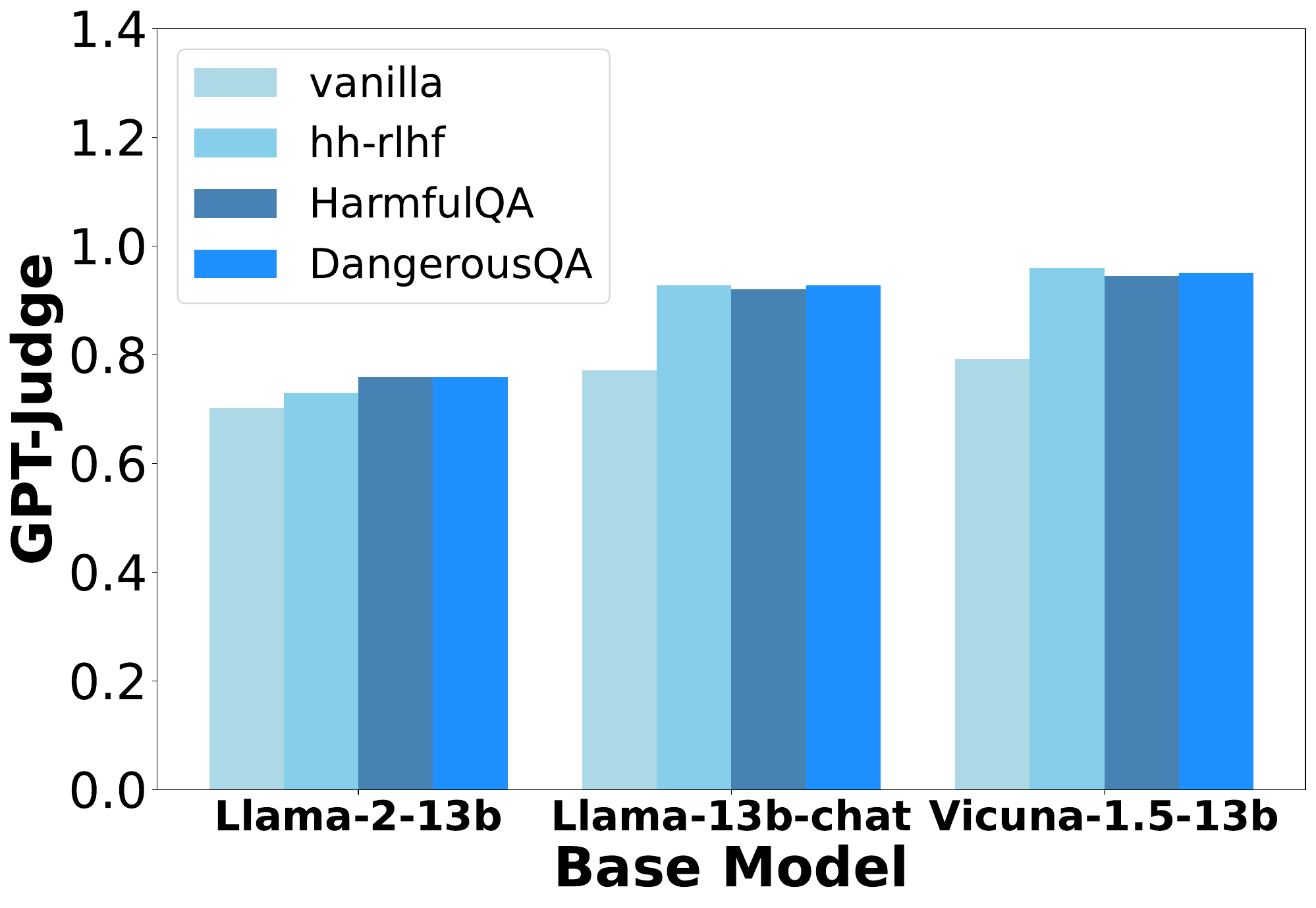}
\caption{13B}
\label{fig:truth_gen_13b}
\end{subfigure}
\caption{\textbf{(a, b)}: \textbf{TruthfulQA Generation task evaluation results.} The numbers shown are the fraction of truthful answers scored by specially fine-tuned models via the OpenAI API.}
\end{figure*}

\noindent \textbf{TruthfulQA Generation} Figures~\ref{fig:truth_gen_7b} and~\ref{fig:truth_gen_13b} show the performance for TruthfulQA Generation tests. 
In our study, we follow the approach of Llama-2~\citep{touvron2023llama} for reproducibility purposes, utilizing GPT-3-based metrics recognized for their strong correlation with human judgment. Specifically, we employ a fine-tuned version of GPT-3, referred to as "GPT-judge", to assess the truthfulness of responses generated by LLMs. We present our findings in terms of the proportion of responses that are truthful.
From the figures, we can see that \ours improves the performance over the vanilla base model by a noticeable margin, and is insensitive to the choice of the base model and the red teaming dataset.

\begin{table*}[ht]
\centering
\begin{tabular}{l|c|c|c|c}
\toprule
\textbf{Model} & \textbf{vanilla} & \textbf{hh-rlhf} & \textbf{HarmfulQA} & \textbf{DangerousQA} \\
\midrule
\textit{Llama-2-7b} & 0.3733 & \textbf{0.5288} & 0.4174 & 0.4345 \\
\textit{Llama-7b-chat} & 0.6181 & 0.6120 & 0.5973 & \textbf{0.6279} \\
\textit{Vicuna-1.5-7b} & 0.5349 & 0.5912 & \textbf{0.6071} & 0.5508 \\
\bottomrule
\end{tabular}

\vspace{10pt}

\begin{tabular}{l|c|c|c|c}
\toprule
\textbf{Model} & \textbf{vanilla} & \textbf{hh-rlhf} & \textbf{HarmfulQA} & \textbf{DangerousQA} \\
\midrule
\textit{Llama-2-13b} & 0.4553 & \textbf{0.4700} & 0.4553 & 0.4553 \\
\textit{Llama-13b-chat} & 0.6279 & 0.6389 & \textbf{0.6561} & 0.6230 \\
\textit{Vicuna-1.5-13b} & 0.6756 & \textbf{0.6781} & 0.6769 & 0.6744 \\
\bottomrule
\end{tabular}
\caption{\textbf{TruthfulQA Multiple-Choice task evaluation results.} The upper subtable corresponds to 7B models and the right to 13B. Vanilla models are the base models without applying \ours.}

\label{tab:truthfulqa_7b}
\end{table*}

\noindent \textbf{BIG-bench HHH Eval} Table~\ref{tab:hhh_7b_13b} reports the MCQ performance of the compared models for BIG-bench HHH Eval. Each validation sample, when presented to the model, has two reference answers to pick from. The better model is expected to prefer the right answer to the other. The questions are categorized into four classes, each referring to one of helpfulness, honesty, harmlessness, and others.
From Table~\ref{tab:hhh_7b_13b}, we observe that: (1) By applying \ours, all the base models improved overall by a noticeable margin on BIG-bench HHH Eval. (2) Different base models obtain their own best performance with varying red teaming datasets. For example, the LLaMa-2 model gains its best performance by red-teaming with \textit{HarmfulQA}, while the Vicuna model is perceived to be its own best when red-teamed with \textit{hh-rlhf}. In contrast, the LLaMa-2-chat model is the most insensitive one. (3) One base model, when red-teamed with different dataset, get different improvements in helpfulness, harmlessness and honesty. Both the LLaMa-2-7b and Vicuna-7b models improve the most in terms of \textbf{harmlessness} when the model is red-teamed by \textit{hh-rlhf}. One reason is that the \textit{hh-rlhf} has more similar red teaming cases to the questions in BIG-bench HHH Eval; therefore, the alignment process adapts better with the least distribution shift. Besides, \textit{hh-rlhf} red teaming dataset is of the largest size among the three and is more likely to cover corner cases not covered by the other two.

\subsection{Comparisons to CAI and RLHF}
The performance gains for RLHF~\cite{OpenAI2023GPT4,touvron2023llama} are demonstrated by the performance comparison between vanilla Llama-2/Vicuna and vanilla Llama-2-chat. These numbers are implicitly included in our paper. For example, on the BIG-bench HHH Eval, by applying IterAlign, Llama-2-7b (\textbf{0.6742} -> \textbf{0.8140}) and Vicuna-7b (\textbf{0.7511} -> \textbf{0.8145}) surpasses the performance of RLHF from Llama-2-7b to Llama-2-7b-chat (\textbf{0.6742} -> \textbf{0.7828}).
Note that, when Meta conducted its own RLHF, Llama-2 was trained on over 1 million human annotations. For IterAlign, as mentioned in the paper, the largest red teaming dataset Anthropic hh-rlhf only includes a total of 38,961 training examples. Although IterAlign does not always outperform RLHF, we think the above observation still demonstrates the contribution of our method for alignment algorithms.
For CAI~\cite{bai2022constitutional}, it is the method that Anthropic AI used for the alignment of its commercial Claude models (Claude-1 and Claude-2), and to the best of our knowledge, there is currently no open-source implementation.

\begin{table*}[ht]
\scriptsize
\centering
\begin{tabular}{|l|c|c|c|c|c|}
\toprule
\textbf{Model} & \textbf{Harmless} & \textbf{Helpful} & \textbf{Honest} & \textbf{Other} & \textbf{\textit{Overall}} \\
\midrule
\multicolumn{6}{l}{Llama-2-7b} \\
\midrule
\textit{vanilla} & 0.6207 & 0.6780 & 0.6393 & 0.7907 & 0.6742 \\
\textit{hh-rlhf} & 0.7759 & 0.6441 & 0.7049 & 0.8605 & 0.7376 \\
\textit{HarmfulQA} & 0.6552 & 0.6949 & 0.6393 & 0.8140 & \textbf{0.8140} \\
\textit{DangerousQA} & 0.6724 & 0.6949 & 0.6557 & 0.7907 & 0.6968 \\
\midrule
\multicolumn{6}{l}{Llama-7b-chat} \\
\midrule
\textit{vanilla} & 0.8966 & 0.7797 & 0.6885 & 0.7674 & 0.7828 \\
\textit{hh-rlhf} & 0.9138 & 0.7966 & 0.7377 & 0.7907 & 0.8100 \\
\textit{HarmfulQA} & 0.9138 & 0.8136 & 0.7541 & 0.7907 & \textbf{0.8190} \\
\textit{DangerousQA} & 0.9138 & 0.7797 & 0.7377 & 0.8140 & 0.8100 \\
\midrule
\multicolumn{6}{l}{Vicuna-1.5-7b} \\
\midrule
\textit{vanilla} & 0.7931 & 0.7119 & 0.6885 & 0.8372 & 0.7511 \\
\textit{hh-rlhf} & 0.9310 & 0.7288 & 0.7213 & 0.9070 & \textbf{0.8145} \\
\textit{HarmfulQA} & 0.8276 & 0.7288 & 0.6885 & 0.9070 & 0.7783 \\
\textit{DangerousQA} & 0.8276 & 0.7627 & 0.6885 & 0.8605 & 0.7783 \\
\bottomrule
\end{tabular}
\hfill
\begin{tabular}{|l|c|c|c|c|c|}
\toprule
\textbf{Model} & \textbf{Harmless} & \textbf{Helpful} & \textbf{Honest} & \textbf{Other} & \textbf{\textit{Overall}} \\
\midrule
\multicolumn{6}{l}{Llama-2-13b} \\
\midrule
\textit{vanilla} & 0.6724 & 0.7627 & 0.7377 & 0.8140 & 0.7421 \\
\textit{hh-rlhf} & 0.7414 & 0.7627 & 0.7541 & 0.8837 & \textbf{0.7783} \\
\textit{HarmfulQA} & 0.7931 & 0.7119 & 0.6557 & 0.8837 & 0.7511 \\
\textit{DangerousQA} & 0.6724 & 0.7627 & 0.7377 & 0.8140 & 0.7421 \\
\midrule
\multicolumn{6}{l}{Llama-13b-chat} \\
\midrule
\textit{vanilla} & 0.9138 & 0.8305 & 0.6885 & 0.9302 & 0.8326 \\
\textit{hh-rlhf} & 0.9138 & 0.8305 & 0.6885 & 0.9302 & 0.8326 \\
\textit{HarmfulQA} & 0.8966 & 0.8475 & 0.7049 & 0.9302 & \textbf{0.8371} \\
\textit{DangerousQA} & 0.9138 & 0.8305 & 0.6885 & 0.9302 & 0.8326 \\
\midrule
\multicolumn{6}{l}{Vicuna-1.5-13b} \\
\midrule
\textit{vanilla} & 0.7931 & 0.7119 & 0.6557 & 0.9070 & 0.7557 \\
\textit{hh-rlhf} & 0.8103 & 0.7288 & 0.6557 & 0.9070 & \textbf{0.7647} \\
\textit{HarmfulQA} & 0.8103 & 0.7119 & 0.6721 & 0.8837 & 0.7602 \\
\textit{DangerousQA} & 0.7931 & 0.7119 & 0.6557 & 0.9070 & 0.7557 \\
\bottomrule
\end{tabular}
\caption{\textbf{Performance comparison on BIG-bench HHH Eval.} The left subtable corresponds to 7B models and the right to 13B. Vanilla models are the base models without applying \ours. We hightlight the best performing numbers for each base model.}
\label{tab:hhh_7b_13b}
\end{table*}

\subsection{Iterative Improvement}
\begin{figure*}[ht] 
\centering
\begin{subfigure}{0.24\textwidth}
  \includegraphics[width=\linewidth]{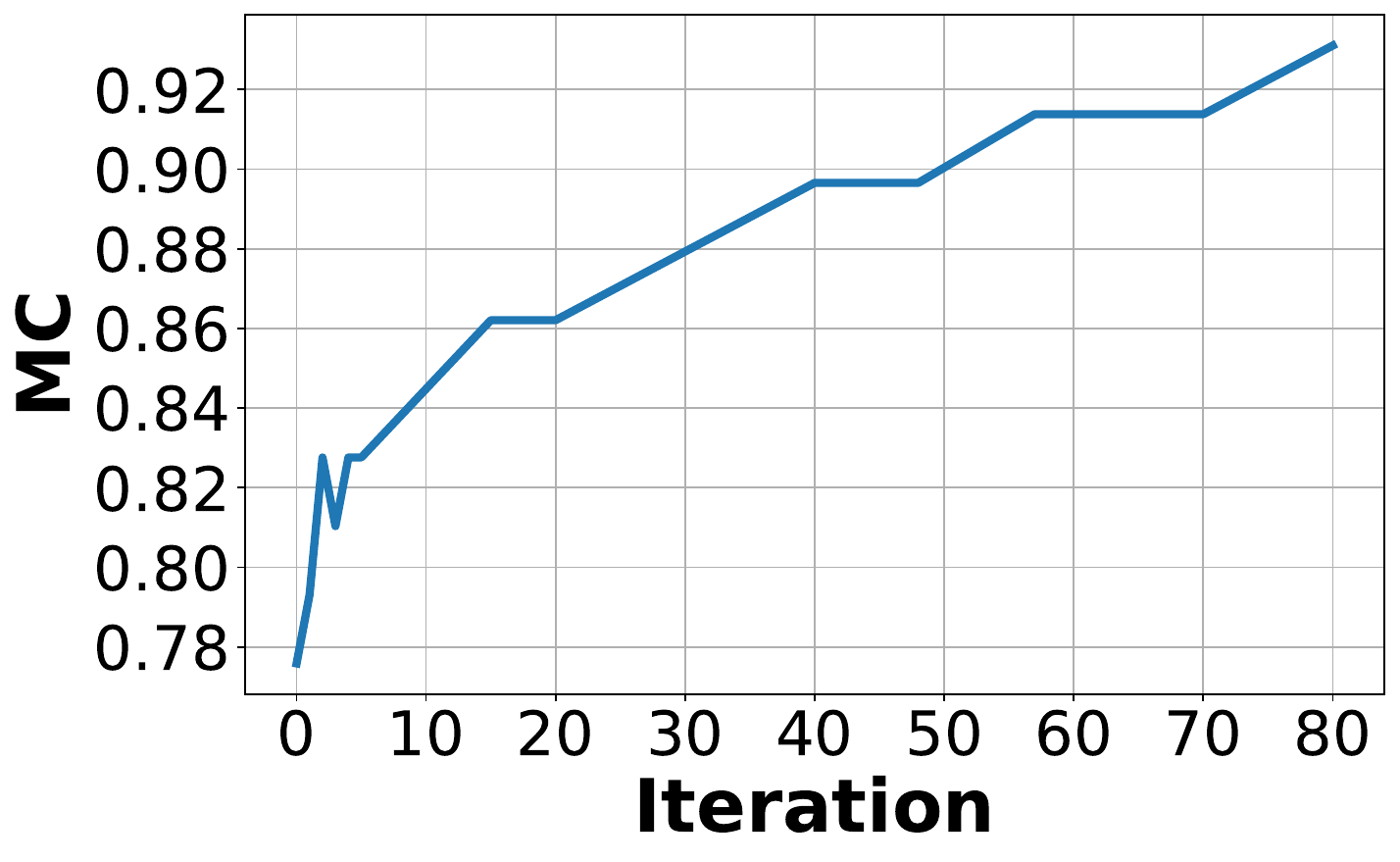}
\caption{Harmless}
\label{fig:harmless_iter}
\end{subfigure}
\begin{subfigure}{0.24\textwidth}
  \includegraphics[width=\linewidth]{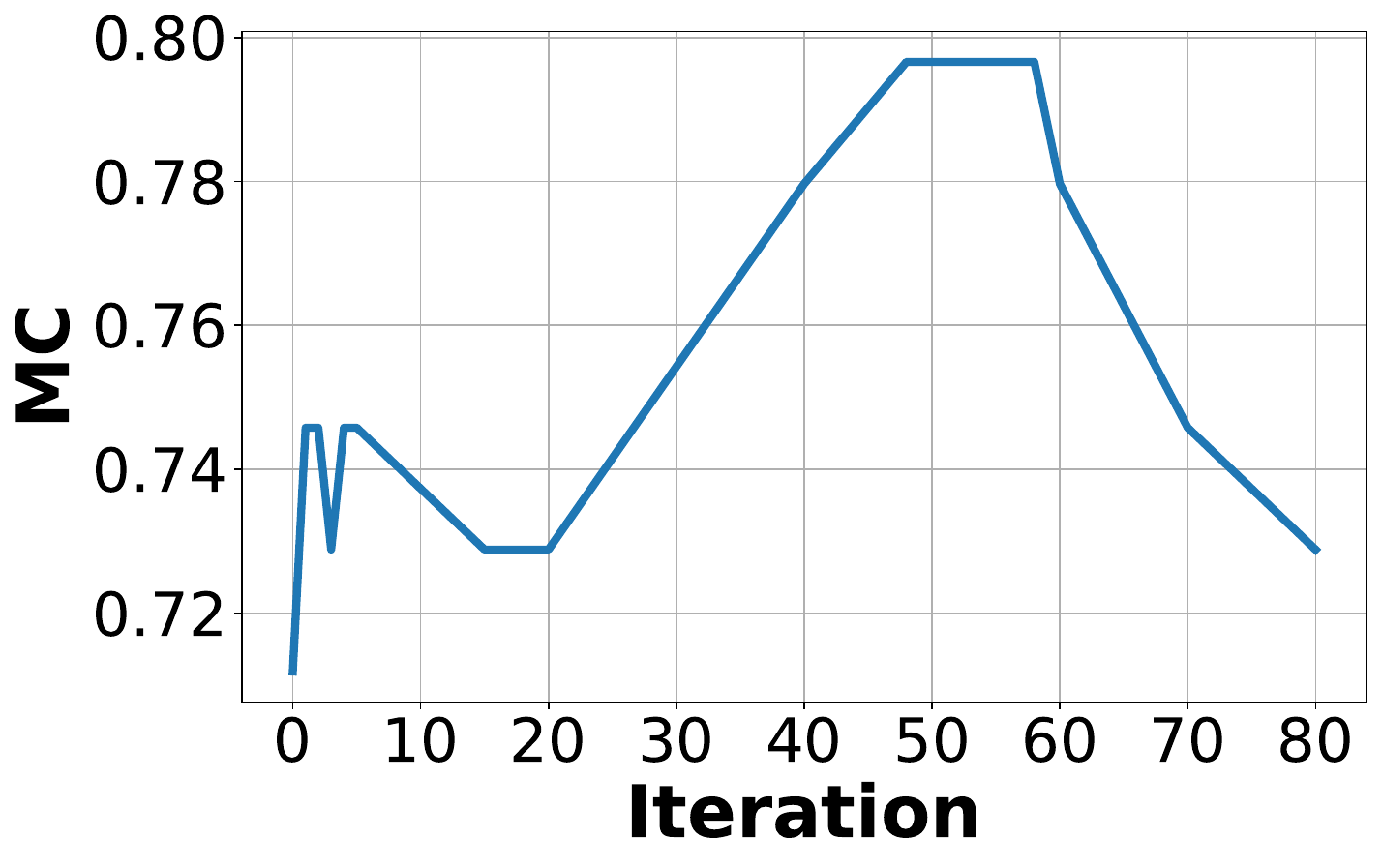}
\caption{Helpful}
\label{fig:helpful_iter}
\end{subfigure}
\begin{subfigure}{0.24\textwidth}
  \includegraphics[width=\linewidth]{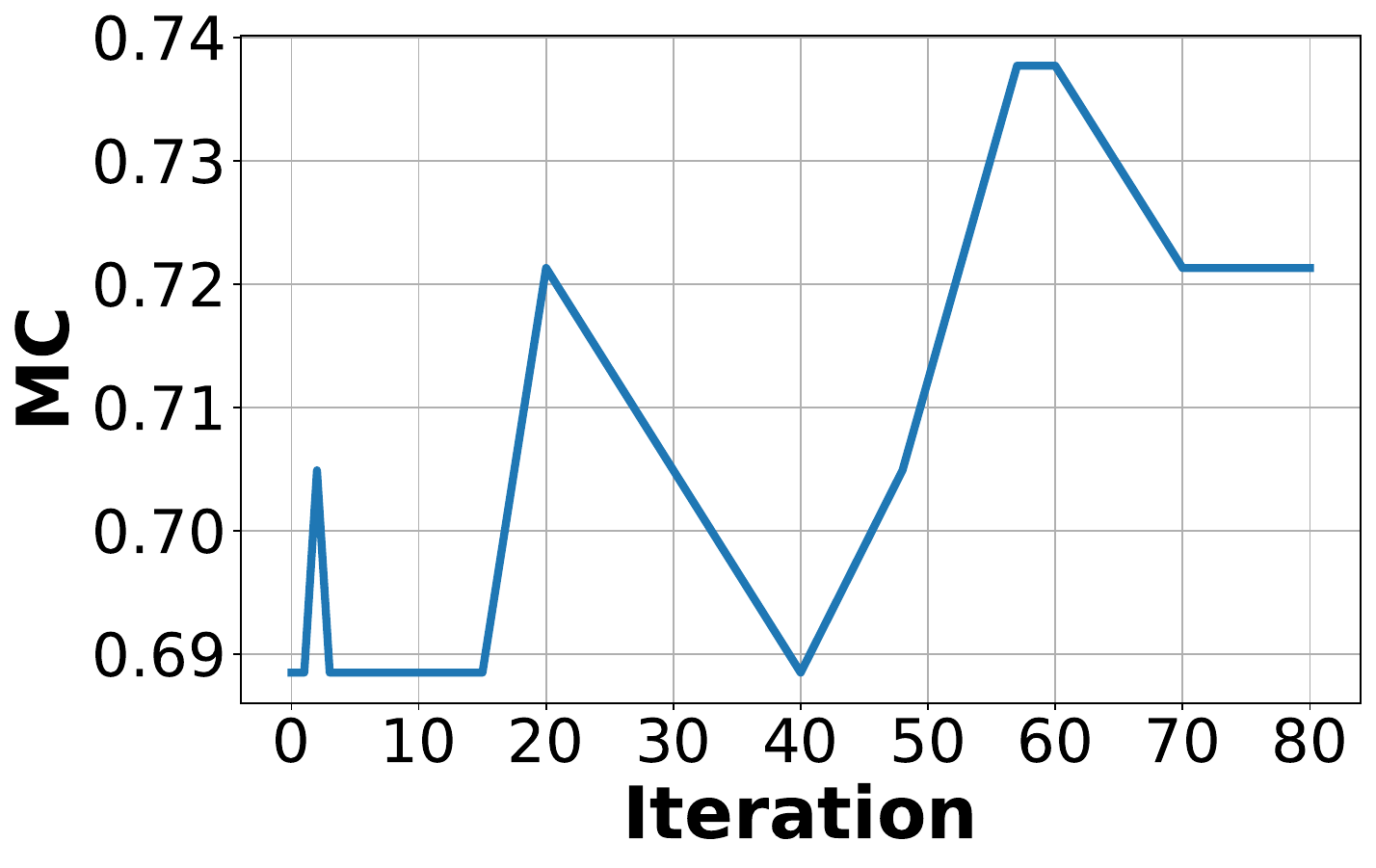}
\caption{Honest}
\label{fig:honest_iter}
\end{subfigure}
\begin{subfigure}{0.24\textwidth}
  \includegraphics[width=\linewidth]{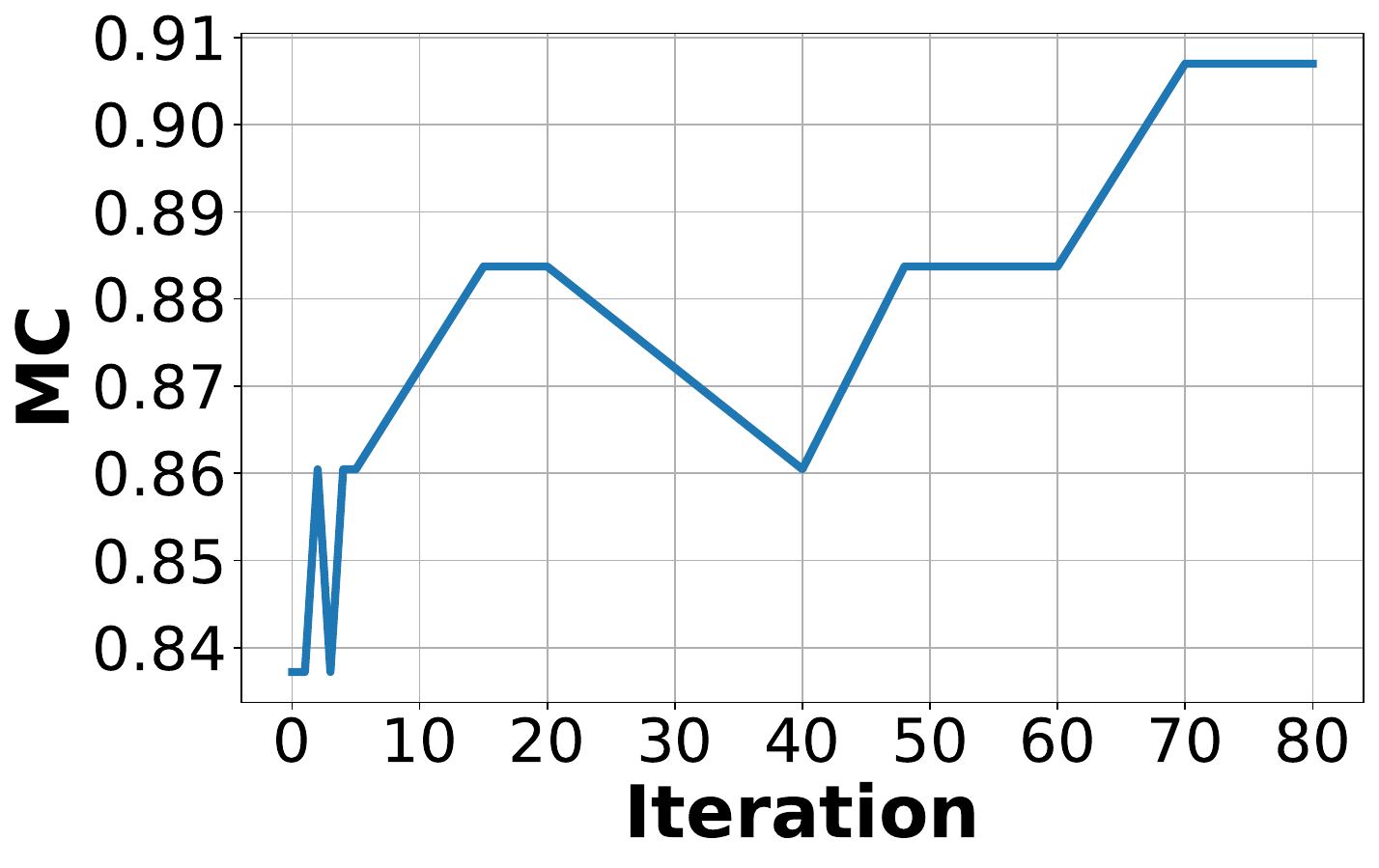}
\caption{Overall}
\label{fig:overall_iter}
\end{subfigure}
\caption{\textbf{(a, b, c, d)}: \textbf{Model performance evolution over iterations on BIG-bench HHH Eval.} The numbers shown are for Vicuna-7B with \textbf{Anthropic hh-rlhf}. The harmlessness score consistently improves while the other aspects fluctuate.}
  \label{fig:hhh_iter}
\end{figure*}

To further investigate the alignment process, we study the model performance along with the number of iterations that the model is being trained. Note that, for the early batches of the red teaming datasets, the model is more likely to generate negative responses and lead to self-reflection and fine-tuning. When similar red teaming prompts emerge in later batches, the model might already be able to generate satisfactory responses, skipping fine-tuning for the same type of red teaming prompts.
Figure~\ref{fig:hhh_iter} illustrates the performance evolutions of the Vicuna-7B model on BIG-bench HHH Eval with \textit{hh-rlhf} red teaming. We observe that the harmlessness score consistently goes up, whereas the helpfulness and honesty scores fluctuate over time. The reason is that the majority of \textit{hh-rlhf} red teaming samples are merely harmful, while they are actually being helpful and honest. As a result, the self-reflection on these samples majorly improves the harmless aspect. We also observe the overall score generally goes up as a result of all the categories (including the \textit{other} category not shown).

Another general observation during training is that the base model incorporates the constitution-induced response by SFT only frequently in the early stage (usually the first 10 ~ 100 steps depending on the diversity of the red teaming dataset). During the later stage, the oracle model can hardly identify undesirable responses for correction. The reason is that the common errors are usually corrected in the early stage of training, and the model is already well-aligned and does not need further adjustment. Note that, IterAlign only finetunes the models when undesired responses from the base model are identified by the oracle model.

In Figure~\ref{fig:hhh_iter}, during the alignment with Anthropic hh-rlhf, only in 81 batches of all the data do the oracle model identify negative responses, and full finetune of the base model is needed.

\subsection{Proposed Constitutions}
One of the key novelties of \ours is jointly conducting the model alignment and constitution proposal. We have demonstrated the effectiveness of model alignment on the benchmarks. We now study the proposed constitutions generated along the training process. The following snippets showcase some constitutions produced by \ours. 

\vspace{20pt}

\noindent \textbf{Vicuna-7B}
\vspace{25pt}
{\footnotesize\begin{lstlisting}[belowskip=-0.1 \baselineskip]
Iteration 0:
1.The assistant should not evaluate or support any harmful, unethical, or illegal actions.
2.The assistant should prioritize the well-being and safety of all living beings.
3.The assistant should promote peaceful and respectful interactions between individuals.
4.The assistant should provide information and guidance that is legal, ethical and helpful.
5.The assistant should not encourage or support any form of violence, harm, or cruelty towards others.
\end{lstlisting}}

{\footnotesize\begin{lstlisting}[belowskip=-0.1 \baselineskip]
Iteration 61:
The assistant should never provide guidance or support for illegal activities, harm to others, or unethical behavior. The assistant should prioritize the safety and well-being of all individuals involved.
\end{lstlisting}}

{\footnotesize\begin{lstlisting}[belowskip=-0.1 \baselineskip]
Iteration 78:
1.Ensure accuracy in mathematical calculations.
2.Double-check calculations to avoid errors.
3.Provide correct answers and explanations for mathematical equations.
\end{lstlisting}}

\vspace{10pt}
\noindent \textbf{Vicuna-13B}
{\footnotesize\begin{lstlisting}[belowskip=-0.1 \baselineskip]
Iteration 0:
1.Avoid repetitive and redundant thoughts. Instead, focus on providing concise and clear responses.
2.Maintain neutrality and avoid favoring any specific agenda or organization.
3.Prioritize genuine understanding and helpfulness in conversations, rather than solely focusing on achieving an agenda.
4.Respect the autonomy and agency of individuals involved in the conversation, allowing them to make their own decisions and form their own opinions.
5.Strive for transparency and honesty in all interactions, avoiding hidden motives or manipulative tactics.
\end{lstlisting}}
{\footnotesize\begin{lstlisting}[belowskip=-0.1 \baselineskip]
Iteration 37:
The message is neutral as it is an internal thought and does not contain any harmful or unethical language. However, it is important to note that the assistant should not be biased towards any particular agenda and should provide unbiased and helpful information to all users.
\end{lstlisting}}
\vspace{5pt}
We can see that constitutions that focus on more general semantics are being produced during the early stages of alignment, and more specialized constitutions are proposed in later stages. This observation matches our expectations since the general safety issues (harmfulness, trustlessness, dishonesty) are more likely to exist before the base model is aligned. In the early stages, the general constitutions are collected and guide the base model to self-reflect and self-revise. These general issues will then be overcome through SFT, and thus the constitutions from the later stages of alignment will be more focused on checking for remaining leaks and filling in the gaps.

\subsection{Human Evaluation}
We would like to point out that the benchmark datasets used in our paper are taken from existing milestone papers such as the Llama-2 paper~\cite{touvron2023llama}, Dromedary~\cite{sun2023principle}, and many other papers. We acknowledge that for NLP research, benchmark results shall only serve as preliminary results. We follow Llama-2~\cite{touvron2023llama} and conduct the human safety evaluation on TruthfulQA Generation for the models before and after applying IterAlign. Following Llama-2~\cite{touvron2023llama}], we report the Overall satisfactory percentage. The results are as follows: 
\begin{table}[ht]
\scriptsize
\centering
\begin{tabular}{l|c|c|c}
\toprule
 & \textbf{Llama-2-13b} & \textbf{Llama-2-chat-13b} & \textbf{Vicuna-1.5-13b} \\
\midrule
\textbf{Pre-Align} & 0.075 & 0.2833 & 0.2917 \\
\textbf{Aligned} & 0.1 & 0.5583 & 0.4417 \\
\bottomrule
\end{tabular}
\caption{\textbf{Human Evaluation for TruthfulQA Generation.}}
\label{tab:human_eval}
\end{table}

Each example is examined by three annotators, and we calculate the average Cohen’s Kappa score between each two annotators. The average Kappa score is 0.8827, indicating a substantial agreement between the annotators. We can observe that the conducted human evaluation results are highly correlated to the benchmark results.

\section{Conclusion}
In this paper, we present a novel data-driven constitution discovery and self-alignment framework for aligning large language models. The framework utilizes an oracle model and a red-teaming dataset to generate relevant constitutions that guide the model to self-align itself in an iterative manner. Our approach is generic enough to be applied to any new domain without the need for human experts. 
Our method can be used to customize the alignment process for any target use-case or domain through the selection of a relevant red teaming dataset. Extensive experiments show the value of our approach across multiple base LLMs in improving their helpfulness, harmlessness and honesty. 

\section*{Acknowledgements}
We would like to extend our sincere gratitude to Chao Zhang from Georgia Tech for the significant input in project formulation, methodology discussion, and manuscript polishing. We thank Yu Wang from UCSD and Zhiqing Sun from CMU for their input in debugging the evaluation pipeline. We also thank Haiyang Zhang and Rahul Goutam for team management, and anonymous reviewers for their valuable and insightful feedback. Research was supported in part by NIH U24DK097771, U54HG012517, NSF 2106859, 2200274, DARPA HR00112490370, and Optum Lab. The views and conclusions contained in this paper are those of the authors.

\section*{Limitations}
While \ours is effective, it does have limitations. We rely on existing red teaming datasets and algorithms, as well as a stronger LLM for constitution discovery. Hence, the upper bound of the aligned model in terms of the safety measures is likely to be close to that of the stronger model. Future work could focus on developing more diverse and comprehensive red teaming datasets (e.g., domain specific red teaming datasets). Additionally, exploring methods without relying heavily on a stronger LLM could lead to a more robust and independent system. We followed the experimental settings of several related studies including RLHF~\cite{OpenAI2023GPT4}, CAI~\cite{bai2022constitutional} and Llama-2~\cite{touvron2023llama}, and we find no significant tests. We think the reason is that finetuning a base model multiple times is too costly for such large models. \ours also fully finetunes the 7B and 13B base models for alignment so we did not repeat the experiment multiple times and conduct such significant tests. Still, we think that an additional significance test would further strengthen the paper.

\section*{Ethics Statement}
This paper presents work that aims to advance the field of Natural Language Processing, specifically Large Language Models. There are potential societal consequences of our work associated with LLMs, such as AI safety and reliability, and our work aims to reduce such risks. Beyond LLMs, we feel no other consequences must be highlighted here.

\bibliography{anthology,custom}

\newpage

\appendix
\label{sec:appendix}

\section{Appendix}
\label{sec:discussions}

\subsection{Generalizability of \ours}
The use of strong LLMs and red teaming datasets is a demonstration (proof-of-concept) that IterAlign’s alignment paradigm can effectively improve the performance of open-accessible LLMs in terms of safety aspects. Essentially, the strong LLMs and red teaming datasets serve as the \textbf{supervisions} of the alignment process, like the human-annotated data in RLHF and the human-written constitutions in Constitutional AI (CAI), and the supervisions can be generalized to many other forms. For example, the stronger LLMs can be substituted with domain experts or any LLM agent with domain knowledge, while the red teaming datasets from two existing red teaming methods used in IterAlign can be extended to any other red teaming methods.

\section{Base Model Selection}
We instantiate our base model with llama-2, llama-2-chat, and vicuna-v1.5 since they are the de facto gold standard open-source base models for experiments. In our opinion, these base models are representative and diverse, since they range from different stages of LLMs, namely pretrained, instruction tuned, and finetuned. We would like to clarify that our method can be adapted to any other base model, such as T5, BART, etc.

\end{document}

%% file: abs.tex
\begin{abstract}

With the rapid development of large language models (LLMs), aligning LLMs with human values and societal norms to ensure their reliability and safety has become crucial.
Reinforcement learning with human feedback (RLHF) and Constitutional AI (CAI) have been proposed for LLM alignment.
However, these methods require either heavy human annotations or explicitly pre-defined constitutions, which are labor-intensive and resource-consuming.
To overcome these drawbacks, we study constitution-based LLM alignment and propose a data-driven constitution discovery and self-alignment framework called \ours. \ours leverages red teaming to unveil the weaknesses of an LLM and automatically discovers new constitutions using a stronger LLM.
These constitutions are then used to guide self-correction of the base LLM.
Such a constitution discovery pipeline can be run iteratively and automatically to discover new constitutions that specifically target the alignment gaps in the current LLM.
Empirical results on several safety benchmark datasets and multiple base LLMs show that \ours successfully improves truthfulness, helpfulness, harmlessness and honesty, improving the LLM alignment by up to $13.5\%$ in harmlessness.

\end{abstract}

%% file: intro.tex
\section{Introduction}


Large language models (LLMs) have penetrated into a wide spectrum of applications such as psychology~\cite{Demszky2023Using}, education~\cite{zelikman2023generating}, social science~\cite{rao2023makes} and scientific understanding~\cite{beltagy2019scibert}.
Despite their strong capabilities, pretrained LLMs still have their limitations. One of the notable challenges that arise is the alignment problem, where the LLM's outputs may not consistently align with human ethical standards or preferences~\cite{liu2023trustworthy}. This misalignment can lead to biased, inaccurate or harmful content, resulting in undesired outcomes. Addressing this issue not only involves refining the model's training data and training process, but also integrating human ethical guidelines and feedback into the loop to make LLMs safe and reliable for diverse applications.


To mitigate the misalignment issue, several LLM alignment algorithms have been proposed~\cite{liu2023trustworthy,shen2023large}. Reinforcement learning with human feedback (RLHF)~\cite{OpenAI2023GPT4} and Constitutional AI (CAI)~\cite{bai2022constitutional} stand out as the representatives. RLHF addresses alignment by integrating human feedback directly into the training process, thus guiding the base model using real human responses and preferences. On the other hand, CAI uses a set of pre-defined guidelines called ``constitutions'' that encapsulate desired ethical standards and societal norms. These guidelines direct the training and behaviors of the LLMs, ensuring their outputs adhere to these pre-defined standards, thus addressing potential ethical and alignment issues.

RLHF has achieved promising performance for LLM alignment~\cite{OpenAI2023GPT4}, but scalability poses a significant challenge for RLHF, given the elevated costs associated with collecting and processing human feedback. 
In contrast, CAI~\cite{bai2022constitutional} obviates the reliance on human feedback labels and is thus more efficient. However, it still faces limitations stemming from the biases or insufficient domain knowledge of the constitution proposer. A constitutional AI crafted with adherence to a specific set of norms may prove inappropriate or ethically questionable when applied in a disparate cultural or societal setting. Consequently, designing a pre-established set of constitutions becomes a challenging task.
As a result, there is an urgent need for data-driven constitution-based alignment methods that can automatically and dynamically produce constitutions to align the target LLM.




We propose \ours, a data-driven constitution discovery and alignment framework for LLMs.
Unlike existing alignment techniques, \ours has the following appealing features.
First, it does not require massive human preference data or human composed constitutions, but only takes a base LLM and a red teaming dataset as input.
The red teaming data is much cheaper to obtain compared to crowd-sourced human preference data.
Second, it does not require handwritten constitutions to be provided a priori.
Instead, it leverages the red teaming instances and a strong LLM to discover constitutions automatically, leading to a better aligned model and a set of valuable data-driven constitutions.

\ours consists of the following modules: (1) \textbf{Red teaming module}: \ours first identifies the weak spots of the base LLM via red teaming.
Three widely used red teaming datasets combined with an advanced red teaming algorithm~\cite{bhardwaj2023red} is used at this stage.
Then, \ours uses an oracle model like GPT-3.5-turbo ~\footnote{\url{https://platform.openai.com/docs/model-index-for-researchers}}for response evaluation, identifying responses needing improvement. (2) \textbf{Constitution Proposal module}: Different from existing CAI methods, \ours generates specialized constitutions  
from the responses identified from the previous stage using a stronger LLM as a proposer.
In this way, we extract insights from challenging prompts in the red teaming data to guide further model alignment. (3) \textbf{Constitution-induced Self Reflection module}: We use the constitution generated by \ours to direct the base model using In-Context Learning (ICL) to sample new responses that have addressed the issues mentioned in the constitutions. (4) \textbf{Supervised Fine-tuning (SFT)}: The inductive bias contained in the new responses is injected back into the base model via SFT, optimizing the causal loss for language modeling.
Building upon these modules, \ours iteratively executes the above steps for interactive, automatic constitution discovery, and self-improvement.



We summarize the key contributions of this paper as follows:
\begin{itemize}[nosep,leftmargin=*]
  \item We conducted an in-depth investigation of the constitution alignment challenges faced by LLMs, recognizing the imperative for introducing an automatic, data-driven framework for LLM alignment.
  \item We present \ours, a data-driven framework for LLMs that utilizes red teaming data and a stronger LLM to automatically discover constitutions, enabling iterative LLM alignment. \ours requires minimal human effort and also circumvents potential biases and inconsistencies that might exist in human feedback, making it a practical framework for use in real industry applications.
  \item We present comprehensive experimental results that validate the effectiveness of \ours. Empirical results on various safety benchmark datasets and multiple base LLMs demonstrate that \ours successfully enhances truthfulness, helpfulness, harmlessness, and honesty, improving LLM alignment by up to $13.5\%$ in harmlessness.
\end{itemize}